\documentclass[conference]{IEEEtran}
\usepackage{epsfig}
\usepackage{graphicx}
\usepackage{amsmath}
\usepackage{amssymb}
\usepackage{rotating}
\usepackage{multirow}
\usepackage{times}

\usepackage[numbers]{natbib}
\usepackage{multicol}
\usepackage[bookmarks=true]{hyperref}

\long\def\comment#1{}

\newcommand{\rtt}[1]{\begin{sideways}{#1}\end{sideways}}

\begin{document}

\title{Reinforcement Learning  for Semantic Segmentation in Indoor Scenes}
%
%%% You will get a Paper-ID when submitting a pdf file to the conference system
%\author{Author Names Omitted for Anonymous Review. Paper-ID [131]}

\author{\authorblockN{Md. Alimoor Reza}
\authorblockA{Department of Computer Science\\
George Mason University\\
Fairfax, VA, USA\\
Email: mreza@gmu.edu}
\and
\authorblockN{Jana Kosecka}
\authorblockA{Department of Computer Science\\
George Mason University\\
Fairfax, VA, USA\\
Email: kosecka@gmu.edu}
}

% avoiding spaces at the end of the author lines is not a problem with
% conference papers because we don't use \thanks or \IEEEmembership

%Homer Simpson\authorrefmark{2},
%Montgomery Scott\authorrefmark{3} and
%Eldon Tyrell\authorrefmark{4}}
%Telephone: (800) 555--1212, Fax: (888) 555--1212}
%\authorblockA{\authorrefmark{4}Tyrell Inc., 123 Replicant Street, Los Angeles, California 90210--4321}}

\maketitle

\begin{abstract}
Future advancements in robot autonomy and sophistication of robotics tasks rest on robust, efficient, and task-dependent semantic understanding of the environment. Semantic segmentation is the problem of simultaneous segmentation and categorization of a partition of sensory data. The majority of current approaches tackle this using multi-class segmentation and labeling in a Conditional Random Field (CRF) framework~\cite{Ren_CVPR2012} or by generating multiple object hypotheses and combining them sequentially~\cite{Banica_2013}.
% The CRF approaches require solving a complex inference problem both in training and testing stage using preset number of semantic labels and large labelled datasets.  
In practical settings, the subset of semantic labels that are needed depend on the task and particular scene and labelling every single pixel is not always necessary. We pursue these observations in developing a more modular and flexible approach to multi-class parsing of RGBD data based on learning strategies for combining independent binary object-vs-background segmentations in place of the usual monolithic multi-label CRF approach.  Parameters for the independent binary segmentation models can be learned very efficiently, and the combination strategy---learned using reinforcement learning---can be set independently and can vary over different tasks and environments.  Accuracy is comparable to state-of-art methods on a subset of the NYU-V2 dataset of indoor scenes~\cite{Silberman_ECCV2012} , while providing additional flexibility and modularity.
\end{abstract}

\IEEEpeerreviewmaketitle

\section{Introduction}
The problem of semantic understanding of indoors environments is central to many service robotic applications where objects need to be sought and recognized. The two most common strategies for tackling the problem of object detection/search is to either train an object category or instance specific detector or  approach the problem as one of semantic segmentation where a partition of the sensory data is labeled with the semantic categories. The later approach can be more powerful as it enables exploitation of various contextual relationships between objects and background or region categories and also between objects themselves. This problem has been studied extensively by the Computer Vision community and evaluated on a variety of benchmark datasets~\cite{Geiger_IJRR2013, Koppula_NIPS2011, Silberman_ECCV2012}.

Most current semantic segmentation approaches use either a multi-label Conditional Random Field (CRF), e.g. \cite{Ren_CVPR2012}, or generate multiple object hypotheses and combining them sequentially, e.g. \cite{Banica_2013,Gupta_CVPR2013,Gupta_ECCV2014}.  Building on the expressive power of these approaches and the availability of large amounts of labeled data, there have been notable recent improvements in accuracy~\cite{Gupta_ECCV2014}.  One appealing aspect of these approaches is that, given enough training data, a system can be trained to perform well on a specific benchmark dataset for semantic segmentation.

\begin{figure}[!htb]
\begin{tabular}{c@{\hspace{0.05cm}}c@{\hspace{0.05cm}}c@{\hspace{0.05cm}}}
\includegraphics[clip,width=0.15\textwidth]{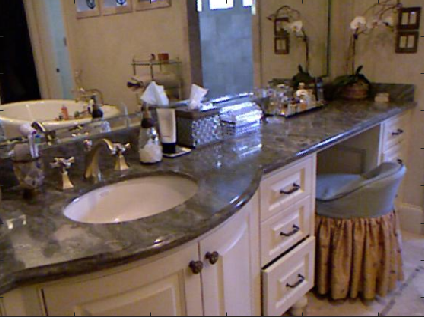}
&
\includegraphics[clip,width=0.15\textwidth]{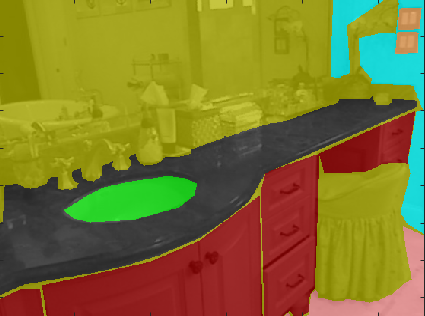}
&
\includegraphics[clip,width=0.15\textwidth]{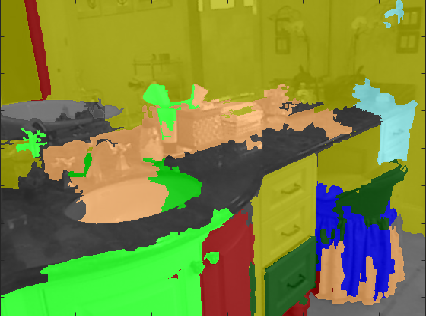}
\end{tabular}
\caption{From left to right, a sample RGB image, corresponding ground truth image, and the semantic segmentation output from our learned policy in a \emph{bathroom} scene.}
\label{fig:vis_of_outputs_nyud}
\end{figure}
The motivation of this paper is to revisit the problem setting from a methodological standpoint.  To this end we observe:  First, depending on the current task or scene, considering all possible labels may not be necessary, while focusing on only task-relevant labels may improve the accuracy of semantic segmentation that is relevant for the task.  Second, as robotics considers wider goals including life-long learning and adaptation, it is less reasonable to assume a known fixed set of labels will be used for parsing. Instead there is a need to represent and acquiring new semantic concepts in a modular and incremental manner while linking them together with additional concepts, e.g. as represented in a large graph by~\cite{Saxena_JSJMK14}.  

Both of these observations touch on one of the core advantages and potential pain-points of approaches that use a multi-label CRF to combine models of local image content based on a variety of features with measures of consistency between nearby labels.  These approaches must optimize the tradeoff between these factors with respect to all of the multiple classes.  This works well when the set of labels is fixed, and the task is semantic segmentation with all those labels, but cannot be easily adapted to consider only a subset of categories relevant for a specific task or environment, or to accommodate a new category.

We propose an alternative approach for semantic understanding of indoors environments, where segmentation and categorization is carried out by learning the sequence of {\em actions} (objects to be recognized and segmented). Central to this approach is that instead of training a large multi-label Conditional Random Field (CRF), we propose learning a number of independent binary object/background segmentations and learning to combine them sequentially, providing flexible modularity as well as some {\em simplification of optimization} in training and inference.  We then formulate the problem of sequential combination of binary object/background segmentations as a Markov Decision Process (MDP).  An additional contribution of the proposed method is the use of variegated superpixels, where large super-pixel regions are formed by robust geometric fitting of planar supporting surfaces and small compact superpixels are using appearance cues.  Contributions include:
\begin{itemize}
  \item reformulation of multi-label CRF based semantic segmentation into modular binary object/background segmentation combined using a learned policy. 
  \item simplified optimization for learning and inference. 
  \item variation in super-pixel formation to accommodate both large planar surfaces and smaller objects.
  \item competitive evaluation on subset of the NYUD V2 dataset.  
\end{itemize}

\section{Related Work}

The presented work is related to different strategies for multi-class semantic segmentation,
for object/background segmentation and context modeling for both RGB images and
RGB-D images. Here we focus on the methodologies used and discuss the suitability of different choices for building modular adaptive life-long learning robotic perceptual systems. 

Baseline approaches for semantic segmentation using RGB-D data were proposed in ~\cite{Silberman_ICCV2011}  where multiple alternatives were considered for the unary and pairwise terms inside a pixel-level Conditional Random Field model (CRF). In the RGB-image only settings, the most successful approaches typically combine local appearance information with a smoothness prior that favors the same label for neighbouring regions (pixels or superpixels)\cite{Gould_ICCV09,Ladicky_ECCV2010}.  More recently, holistic scene representations were proposed using a graphical model framework~\cite{Lin_ICCV2013}. In ~\cite{Ren_CVPR2012, Koppula_NIPS2011} the authors used superpixel hierarchies endowed by features extracted from the entire path of the segmentation tree. 
In later work, excellent results were achieved by considering bottom up unsupervised segmentation and boundary detection to generate and classify candidate regions~\cite{Gupta_CVPR2013}.  More recent approaches achieved superior results, by using deep convolutional networks for feature extraction, trained jointly or independently with CRF models used to capture the context and spatial relationships between objects ~\cite{Ladicky_ECCV2010, ShottonCVPR08}. 
The above mentioned approaches exhaustively consider all semantic labels and either train 
a multi-label Conditional Random Field (CRF) or multiple one-vs-all classifiers evaluated on the regions. In ~\cite{Couprie_ICLR2013}, the authors also bypass the complex feature computation and segmentation stage and use convolutional networks for semantic segmentation. Hypotheses for the indoor scene labelling task are then evaluated using a superpixel-based over-segmentation. \\

% In case of RGB-D data several semantic segmentation methods have been developed mostly for indoor environments. In \cite{Koppula_NIPS2011}, authors highlighted the need for efficiency of the final inference and used up to 17 object classes and moderate variations in the scenes to evaluate their results. They were able to exploit stronger appearance and contextual cues capturing the relationships between objects in 3D. More recently, more comprehensive experiments have been conducted on larger NYU-V2 RGB-D dataset introduced in~\cite{Silberman_ICCV2011}.

% A different  strategy is followed by \cite{Gupta_CVPR2013,Gupta_ECCV2014} where the authors used the available depth information to improve the initial segmentation, followed by classification of obtained segments. \\

\noindent
The problem of binary object/background classification has been also tackled with the commonly used sliding window processing pipelines, with window pruning strategies and features adapted to RGB-D data. The commonly used approaches include~\cite{Pepik_ECCV2012, Ye_EECS2013}. The bounding boxes generated by the sliding windows approaches, while efficient, provide only poor segmentation and are often suitable for objects whose shape can be well approximated by a bounding box.  Alternative strategies for object detection and semantic segmentation, proceed with mechanisms for generating object proposals from elementary segments using low-level grouping cues of color and texture. This approach has been pursued in the context of generic object detection followed by recognition in the absence of depth data by \cite{Sande_ICCV2011}. Strategies for generating and ranking multiple object proposals have been also explored by \cite{Carreira_PAMI2012}. Discriminatively-trained part-based models, which incorporated some context information have been introduced recently~\cite{Mottaghi_CVPR2014}. \\

A bodies of work explored Reinforcement Learning techniques for robotics and image understanding problems \cite{Karayev_NIPS2012}\cite{Kwok_IROS2004}. Karayev and colleagues \cite{Karayev_NIPS2012} learn a policy that optimizes object detections in a time restricted manner. Kwok et al.\cite{Kwok_IROS2004} introduced an augmented MDP formulation for the task of picking the best sensing strategies in a robotics experiment and utilizes least square policy iteration to solve for the optimal policy.
%{\em JK add some sequential composition strategies of Simchinescu}

\section{Proposed Approach}
We start by learning binary object/background segmentation models  for each \emph{object} category.  While these do use a CRF to integrate appearance and label consistency cues, these are not mulit-label CRFs and only consider a single category and the background.  A the resulting object/background segmentations may overlap, they need to be combined into a single pixel-level labeling. The simplest strategy of taking the most confident prediction for the overlapping regions is not ideal as it requires calibrating the categories vs each other, something we explicitly avoid, and often prefers large objects of interest which can be detected with high confidence compared to smaller objects.  Instead, we adopt a sequential strategy for combing the binary segmentations, where the strategy is found using reinforcement learning.
This approach is favorable compared to a fixed ordering, which naturally gives priority to large or frequent categories due their contribution to the overall pixel level accuracy performance measure, and can dynamically adjust based on each image.  

In the following sections we describe the ingredients of our approach starting with learning and inference for generating binary object/background segmentation, and then the sequential combination strategy and learning the optimal sequencing of object background segmentations in reinforcement learning framework. 
 
\subsection{Object/background segmentation} 

\begin{figure}[t]
\centering
% chair table pair
\includegraphics[clip,width=0.20\textwidth]{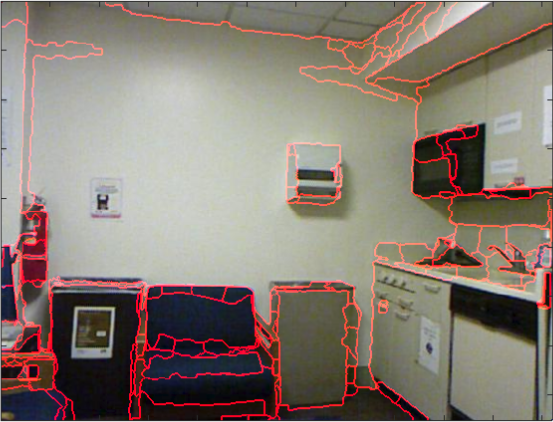}
\hspace{0.1cm}
\includegraphics[clip,width=0.20\textwidth]{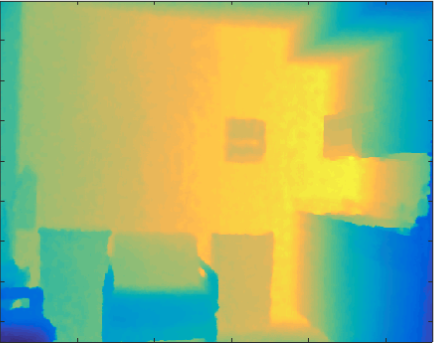}
\hspace{0.1cm}
\caption{Our superpixel image (bounded by red lines) and the depth image from NYUD V2. Notice how our superpixels capture the large planar surfaces.}
\label{fig:sp}
\end{figure}

Motivated by the observation that indoor scenes contain many planar structures, we segment an image into regions of planar and non-planar surfaces. Our framework starts by identifying the dominant planar surfaces using a RANSAC based approach by sampling 3D point triplets from the depth image. The remaining regions not explained by the planar surfaces are represented by compact SLIC superpixels~\cite{Achanta_PAMI2012} generated from the RGB image. Figure~\ref{fig:sp} shows our superpixels, where the region boundaries are marked in red. \paragraph{Conditional Random Fields}
\label{section:crf}
We formulate the binary object/background segmentation in a Conditional Random Field (CRF) framework. The conditional distribution $p(\bf{x}|{\bf{z}})$, is a log-linear combination of feature functions or potentials as shown in \ref{eq_2}. By maximizing the conditional likelihood, we can learn the weights $w_i$'s from a set of labeled training examples. Final prediction for the hidden random variables is found as the $\emph{maximum a-posteriori (MAP)}$ solution ${\bf{x}}$ for which the conditional probability $p(\bf{x}|\bf{z})$ is maximum.
%\begin{equation}
%\begin{split}
%p(\mathbf{x}|\mathbf{z})=\frac{1}{Z(\mathbf{z})}\exp (w_d\sum_{i}^{} \theta_d(x_i,\mathbf{z})\\
%+\sum_{\emph{t}}^{}\sum_{(j,k)\in\xi}^{}w_\emph{t}\theta_{\emph{t}} (x_j,x_k,\mathbf{z}))
%\end{split}
%\label{eq_2}
%\end{equation}
\begin{equation}
\begin{split}
p(\mathbf{x}|\mathbf{z})=\frac{1}{Z(\mathbf{z})}\exp (w_1\sum_{i}^{V} \theta_d(x_i,\mathbf{z})+w_2\\
\sum_{(i,j)}^{E}\theta _{pc} (x_i,x_j,\mathbf{z})+w_3\sum_{(i,j)}^{E}\theta _{px} (x_i,x_j,\mathbf{z}))
\end{split}
\label{eq_2}
\end{equation}
The hidden random variables $\bf{x}$=$\{x_1, x_2, ..., x_{\left | V \right |}\}$ correspond to the nodes in the CRF graph,  $\bf{z}$  refers to the vector of observation variables, and $Z(\mathbf{z})$ is the partition function. Each hidden random variable can take one of two discrete values: ${x_i}$ =  {\em \{object, background\}}, where $object\in$  {\em \{bed, sofa, ... ,  wall\} } in indoor scenes experiment.  Note that the object here simply refers to one of the semantic labels available in the dataset. 
% The CRF graph is denoted by $G=\{V,E\}$, where $V$, $E$ are the graph nodes and edges respectively.\\
The {\em unary feature function} associated with each node, $x_{i}$, comes from the output probability $p(x_i|\bf{z})$ of a AdaBoost classifier. 
\begin{equation}
\theta_d(x_i,\mathbf{z}) = -log(p(x_i|\bf{z}))
\end{equation} 
where the observations $\mathbf{z}$ are computed for each superpixel $i$ using a subset of features described in Section \ref{features}. 
The {\em pairwise functions} are computed for every edge ($j,k$) $\in$ $E$ in the CRF graph. We penalize the labelings of two adjacent superpixels according to their differences in one of two aspects $\in \{color, spatial\}$.  Color terms penalized color differences, while the spatial term penalizes the differences between the mean depth associated with each superpixel as detailed in \cite{cadena_2014ijrr}. These pairwise functions ensure the smoothness in the final labeling process for superpixels. We follow similar CRF graph construction method, learning and inference technique as detailed in \cite{cadena_2014ijrr}.
\comment{
The color pairwise term $\theta_{\emph{pc}}(x_i,x_j,\mathbf{z})$ is defined using the following equation:
\begin{equation}
\theta _{pc}(x_i,x_j,\mathbf{z})=\left\{\begin{matrix}
1- exp(-\left \| {\bf c_i}-{\bf c_j} \right \|)& , & l_i=l_j\\ 
exp(-\left \| {\bf c_i}-{\bf c_j} \right \|)& ,& l_i\neq l_j 
\end{matrix}\right.
\end{equation}
where $\left \|{\bf c_i}-{\bf c_j} \right \|$ is the $L_{2}$-norm of the difference between two superpixels color. ${\bf c_{i}}$, ${\bf c_{j}}$ correspond to the  3-element mean color vectors in the \emph{Lab} space. The spatial pairwise term, $\theta _{px}(x_i,x_j,\mathbf{z})$, defined similarly from the average 3D positions vector ${\bf t_{i}}$, ${\bf t_{j}}$ of the point clouds associated with superpixels $i, j$ respectively and $l_i, l_j$ corresponds to the labels associated to the superpixels $i,j$.
\begin{equation}
\theta _{px}(x_i,x_j,\mathbf{z})=\left\{\begin{matrix}
1- exp(-\left \| {\bf t_i}-{\bf t_j} \right \|)& , & l_i=l_j\\ 
exp(-\left \| {\bf t_i}-{\bf t_j} \right \|)& ,& l_i\neq l_j 
\end{matrix}\right.
\end{equation}
}

\subsection{Features}
\label{features}
We  use a set of rich and discriminative features to characterize the superpixels capturing both local appearance and geometry and global statistics of the scene.  We exploit a subset of {\em geometric} and generic features proposed in~\cite{Gupta_CVPR2013}, computed over superpixels.   Appearance is encoded using the traditional representation of {\em color} and {\em texture} features introduced in ~\cite{Uijlings_IJCV2013} and ~\cite{Hoiem_IJCV2007}. The color distribution of a superpixel is represented by a fixed histogram of color in HSV color space and texture histogram is formed from a set of oriented  Gaussian derivatives in a fixed set of orientations. %The texture is represented using oriented filters. Gaussian derivatives are taken in each of the eight orientations for each color channel in HSV and a 10 bin histogram is extracted for each orientation in a particular channel. 
% The concatenation of these histograms over the three separate channels results in a histogram of 240 bins that represents our texture feature. 
 % Similar representation is used in the work of Uijlings et al. \cite{Uijlings_IJCV2013}, but on a different context of object detection.
% We also adopt a set of efficient features for {\em occlusion} reasoning as introduced in~\cite{Cadena_IJRR}.
The geometric features computed over these 3D points associated with each superpixel  are described in detail in \cite{cadena_2014ijrr}. 
We also adopted a set of perspective features computed from the the statistics over the straight lines and their intersections and other vanishing point statistics as described in~\cite{Hoiem_IJCV2007}.

\subsection{Unary Terms}
We learn the AdaBoost classifier in a one-vs-all fashion for a particular object such as \emph{bed}, \emph{table}, \emph{sofa} and others. We utilize the logistic regression version of AdaBoost introduced in~\cite{Hoiem_IJCV2007}. We apply sigmoid function on the predicted output of the AdaBoost, a confidence measure, to get a probabilistic output of a superpixel being a label of the object under consideration. 
%We train our AdaBoost classifier in a \emph{one-vs-all} fashion for all semantic labels.  We consider only those images in the training set that contain the instance of \emph{object}. We label an example, a high dimensional feature representation of a superpixel, as an instance of the positive class if more than 80\% of the constituent pixels' labels belong to the positive class in an image. We label the remaining examples as instances of the negative \emph{background} class. 
We maintain approximately equal proportions of positive and negative instances during training to prevent over-fitting. We use a {\em negative mining}  strategy that utilizes the co-occurrence statistics to select important instances from a large pool of negative instances. The important negative instances are sampled according to the distribution of \emph{object}'s co-occurrence with other objects in the training images. % For example, Figure \ref{co_oc_img_nyud}a) shows the co-occurrence matrix of pairwise presence of objects. Figure \ref{co_oc_img_nyud}b) shows the co-occurrence distribution for \emph{bookshelf}, which indicates  \emph{books}  co-occur most frequently with \emph{bookshelf} in NYUD-V2 dataset. 
We select negative samples from a particular object class in proportion to the object's co-occurrence value in the distribution. For example during training for \emph{books}, we will sample more examples from \emph{bookshelf} class as negative instances compared to other non-frequently co-occurring objects e.g., \emph{toilet}. Our trained model with this negative mining strategy shows improved performance in terms of overall accuracy compared to random selection of negative instances.

\comment{
\begin{figure}
\includegraphics[width=1.5in]{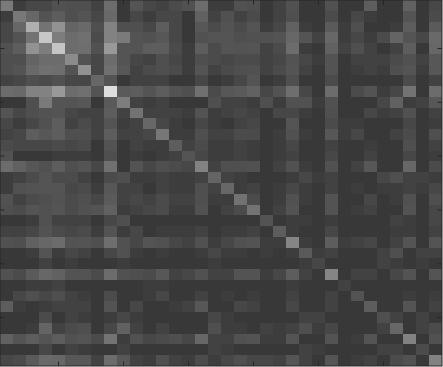} \includegraphics[width=1.5in]{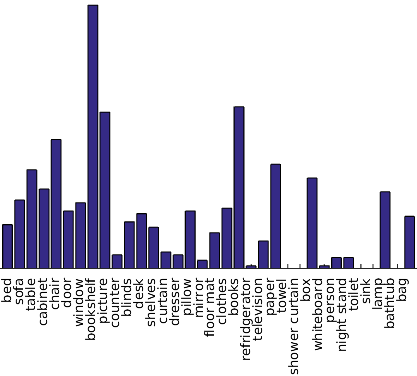}
 \caption{a) 34x34 object class co-ocurrence matrix in NYUD-V2~\cite{Silberman_ECCV2012} (white entries denoting higher co-occurrence values); b) corresponding bookshelf row where two highest peaks belong to book and bookshelf categories.}
\label{co_oc_img_nyud}
\end{figure}
}

\subsection{Sequential combination of binary segmentations}
Traditional approaches for semantic segmentation look for all possible labels in the scene \cite{Yao_CVPR2012, Gupta_CVPR2013}. They either approach the problem by training a multi-class CRF or one-vs-all classifiers for all semantic labels and computing the confidences for all the semantic labels. We next describe our sequential object selection mechanism and object selection policy learning using reinforcement learning.  The resulting policy dynamically selects a sequence of objects, completely based on the image content.

%selects objects in a scene dynamically . 
% \subsubsection{Sequential combination of binary segmentations} 

\begin{figure}
%\centering
\includegraphics[width=1.0in]{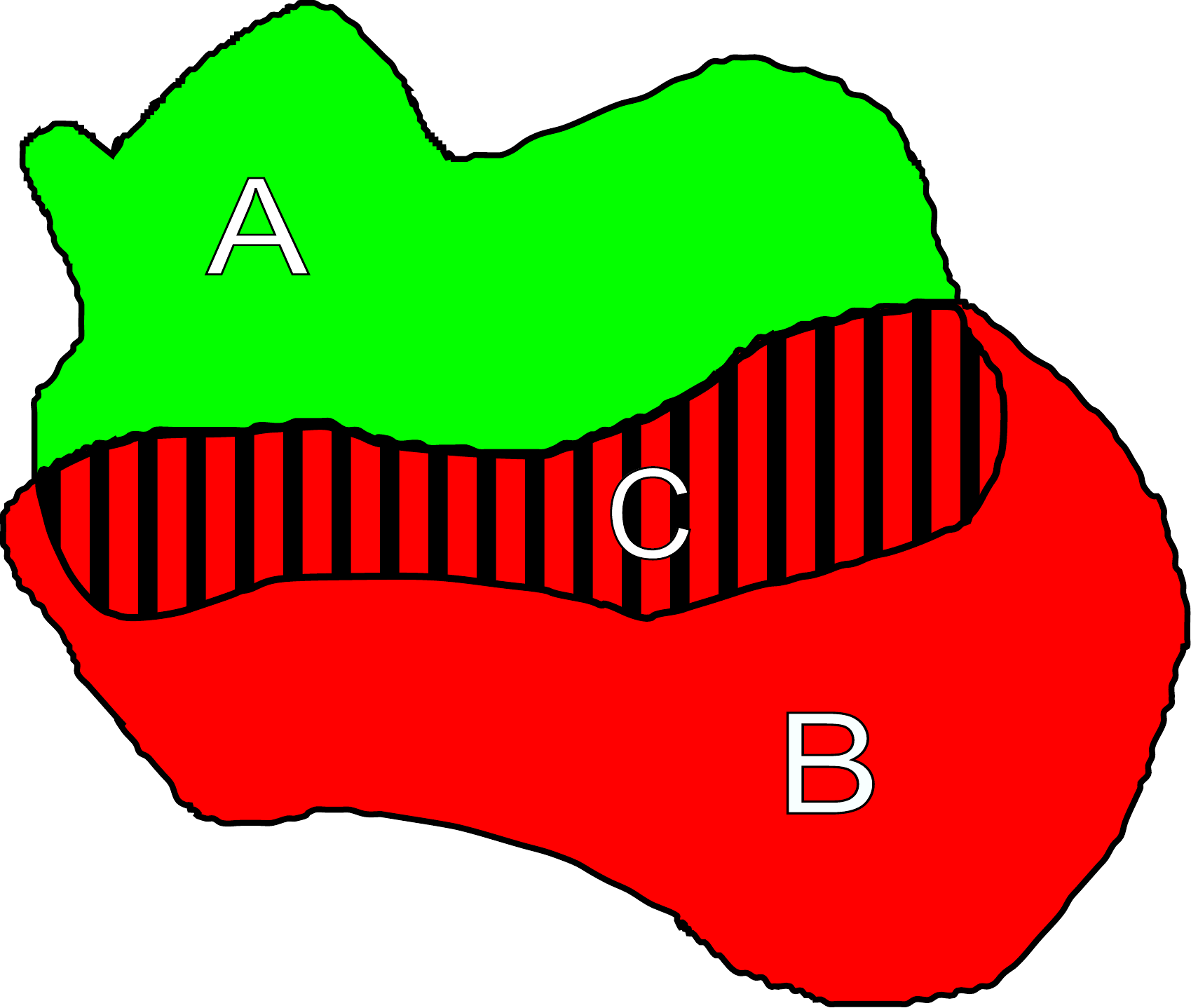}
\includegraphics[width=2.5in]{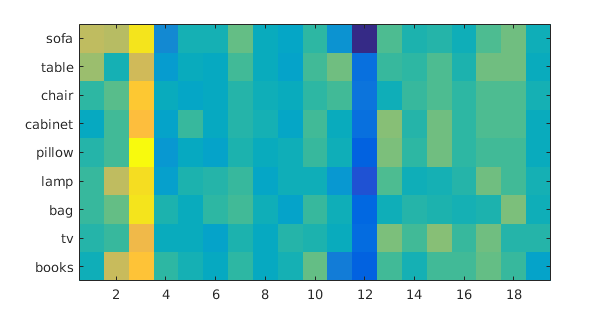}
\caption{ a) Conflict resolution of intersected region between two competing object segments discussed in \ref{sec:conflict}. b) Learned weight vector ${\bf w_{\pi}}$ corresponds to the learned policy for the \emph{living room} the 3-Fold cross validation experiment. The yellowish entries denote positive values where as bluish entries denote negative values. y-axis shows the actions and each row denotes the weights for each action blocks.}
\label{fig:conflict}
\end{figure}

\label{sec:conflict}
We generate the binary object/background segmentations %We find the most frequently occurring \emph{K} objects 
for different objects in each scene and sequentially combine their predicted segmentation masks. To resolve the conflict between an overlapping region we utilize a heuristic that considers overlap ratios of two competing segments. For example, in the sequential combination process of the \emph{bedroom} scene, object \emph{bed} is considered first followed by object \emph{pillow}. \emph{C} is the overlapped region between segments \emph{$B$} and \emph{$A$}, which have different labels as shown in Figure~\ref{fig:conflict}. The regions are broken into the following $A=C+A{'}$ and $B=C+B{'}$, where $A{'}$ and $B{'}$ are the non-intersected portion of segments $A$ and $B$ respectively. %Segments \emph{A} and \emph{B} come from label of object \emph{a} and \emph{b}
Prior to considering \emph{pillow} object in the sequential process, the overlapped part \emph{C} was assigned the label \emph{$C_{bed}$} along with \emph{$A{'}_{bed}$}. The non-overlapping part of \emph{pillow} has label  \emph{$B{'}_{pillow}$}. We consider the segment-size-ratio heuristic, which prefers small-sized segments that can take place on-top-of larger segments e.g., \emph{pillow} can lie on-top-of \emph{bed}. We compute the ratios $\frac{C}{A}$, and $\frac{C}{B}$ then assign \emph{C} to labels of the segment that has the larger ratio value.
For an image, we sequentially combine obtained binary segmentations from three background classes in the order of \emph{wall, floor}, and \emph{ceiling} irrespective of the scenes we consider, then follow it up with sequential selection of objects that is given by a policy we learn using reinforcement learning.

\subsubsection{Markov Decision Processes}
In this section section we will review the reinforcement learning framework based on least squares policy iteration introduced in~\cite{Lagoudakis_JMLR2003}.  The general framework of learning control policies (sequences of actions) based on experience and rewards is that on Markov Decision Processes (MDP). MDP models the agent acting in the environment as a tuple  $\{ S,A,T,R\}$, where 
{\em S} is a set of states, {\em A} finite set of actions, $T(s, a, s')$ is the transition model which describes the probability of of ending up in state $s'$ after executing an action $a$ in state $s$
and $R(s, a, s')$ is a reward obtained in state $s$ when agent executes action $a$ and transitions to $s'$.  The goal of solving MDP is to find a policy $\pi: S \rightarrow A$ for choosing actions in states such that cumulative future reward is maximized.  If parameters $T$ and $R$ are known 
then optimal control policy can be computed efficiently using value iteration. In case the parameters of MDP are now known, the reinforcement learning can be used to learn the optimal policy. 
One set of RL techniques aims at learning the model parameters $T$ and $R$  and then proceeds with standard value iteration, the other {\em model-free} techniques learn the policy directly.
For our problem we adopt the {\em model-free} approach and apply least squared policy iteration introduced in~\cite{Lagoudakis_JMLR2003} and also used in \cite{Kwok_IROS2004} and \cite{Karayev_NIPS2012} in the context of robotics and image understanding problems. 

\subsubsection{Model} 
In our case the state 
$\emph{s} \in S$  consists of our current estimate of the distribution of the objects in the scene $P({B}) = \{P(B_{1}), P(B_{2}), ..., P(B_{K})\}$, where $P(B_{i})$ is the probability of an object class-$i$ is present in the image. Additionally, we also keep track of objects that are already taken into consideration into a observation variable ${O}$.  A set of actions $\emph{a}_i \in {A}$, where action $\emph{a}_i$ corresponds to the selection of object/background segmentation  for object $i$ in the sequential semantic segmentation.  Our reward function $R$ is defined in terms of pixel-wise frequency weighted Jaccard Index 
computed over the set of actions taken at any stage of an episode. 
{\em Jaccard Index}  $JI_{i}$ is the intersection-over-union ratio for the object $i$'s predicted segmentation and the ground truth segmentation. In the training stage we consider each image as an episode. Length of an episode is the maximum number of actions in our action set $A$.  Lets assume that in the current episode, starting with action $a_{1}$ we reached to position $\emph{k}$ of the episode by taking $a_{k}$. MDP state maintains actions taken so far in observation variable ${O}$. We compute the current reward $R(s,a_{k},s')$ in state $s$ at the episode as follows:
\begin{equation}
\begin{split}
R(s,a_{k},s')=w_1*JI_{a_1}+w_2*JI_{a_2}+ ... + w_k*JI_{a_k}+r_{bg}\\             
\end{split}
\label{eq_9}
\nonumber
\end{equation} 
\begin{equation}
\begin{split}
r_{bg}=w_{wall}*JI_{wall}+w_{floor}*JI_{ceiling}                 
\end{split}
\label{eq_10}
\end{equation}
$w_{i}$ are the ratio of the pixels predicted by the segmented object $i$ to the total pixels in the image. The $r_{bg}$ is the \emph{pixel-wise frequency weighted Jaccard Index} score of the three fixed background classes. 
%We find it more suitable to prefer the \emph{pixel-wise frequency weighted Jaccard Index} over \emph{average Jaccard Index}.
Our dynamic policy finds the state-action value function $Q^{\pi }(s,a)$, which estimates a real-valued score to each state-action pair as defined by equation \ref{eq_3}.
\begin{equation}
\begin{split}
Q^{\pi }(s,a)=E[R(s,a,s^{'})+\gamma Q^{\pi }(s^{'},\pi(s^{'}))]
\end{split}
\label{eq_3}
\end{equation}
The policy is defined on the $Q^{\pi }(s,a)$ as defined by equation \ref{eq_4}. Our state space is continuous and intractable, hence we can featurize the policy function with an approximation function with features $\psi(s,a)$ and a set of linear weights $w_{\pi}^{T}$. 
\begin{equation}
\begin{split}
\pi(s)=\arg\max_a Q^{\pi }(s,a)\\
 =\arg\max_a w_{\pi}^{T}\psi(s,a)
\end{split}
\label{eq_4}
\end{equation} 

\subsubsection{State Featurization of MDP}
Our MDP is not fully observable hence we follow a pursuit of Augmented MDP\cite{Karayev_NIPS2012}\cite{Kwok_IROS2004} which includes uncertainty variables into the state featurization. $\psi(s,a)$ is defined by following a block-featurization approach where features specific to each action are inserted into a specific block. All blocks are concatenated one after another to give us a fixed length feature vector. Each block in $\psi(s,a)$ has fixed order of entries. \begin{equation}
\begin{split}
\psi_{a}(s,a)=\{P(B_{prior}^{a}),P(B_{1}|{\bf O}),..., P(B_{K}|{\bf O},\\
 U(B_{1}|{\bf O}),..., U(B_{K}|{\bf O})\}
\end{split}
\label{eq_5}
\end{equation}
Here $P(B_{prior}^{a})$ is the prior probability of the object $a$ in the image. $P(B_{i}|{\bf O})$ are the object probabilities conditioned on the current set of observation. $U(B_{i}|{\bf O})$ are the uncertainties conditioned on the observations, which is computed as the Shanon-Entropy measure from the $P(B_{i}|{\bf O})$. Each block is of size 1+2\emph{K}, where \emph{K} is the size of a single block feature as mentioned above. If size of the Actions set is $|A|$, then our featurization function $\psi(s,a)$ has size of $|A|$(1+2\emph{K}). All but the selected action block is zero in this featurization approach.
\subsubsection{Modeling the Featurization Components}
The block-featurization of action \emph{a} is computed as follows.
The prior term $P(B_{prior}^{a})$  is computed from the MAP probability of the binary segmentation for object \emph{a}. %We compute the connected components of the binary segmentation mask for object $a$. The connected components larger than 50 pixels are retained for computing the $P(B_{prior}^{a})$. For each of these connected components, we compute the average MAP values of all the pixels as $P(B^{a}_{{prior}_{j}})$ . The final $P(B_{prior}^{a})$ is computed as follows:
%\begin{equation}
%P(B_{prior}^{a}) = \frac{\sum_{j}^{CC}P(B^{a}_{{prior}_{j}})}{\left |CC \right |}
%\label{eq_13}
%\end{equation}\\
The observation terms updates the belief-state each time an action is performed. The selection of object changes the our belief about the presence or absence of the object in the image. In order to capture this information, we update the observation terms $P(B_{i}|{\bf O})$ for $\emph{i}$-th object if action $a_{i}$ is in the observation vector ${\bf O}$ by the following equation:
\begin{equation}
P(B_{i}|{\bf O}) = \max \{P(B_{i}|CC^{1}),..., P(B_{i}|CC^{M})\}
\label{eq_13}
\end{equation}
Each $CC^{j}$ term is the connected components from the segmentation mask of object $i$. Each $P(B_{i}|CC^{j})$  is computed from the output of a classifier referred as RUSBoost ~\cite{SeiffertKHN2010}. We train the classifier from the ground truth segmentation mask for each object from the training images. We used the same features as described in Section\ref{features}. 
$P(B_{i}|{\bf O})$ retained the value $P(B_{i})$ if action $a_{i}$ is not the observation vector ${\bf O}$
\subsubsection{Reinforcement Learning for Policy}
The goal is to learn a policy, which maximizes the state-action value function, $Q^{\pi }(s,a)$ defined in Equation~\ref{eq_3}. $Q^{\pi }(s,a)$ is defined recursively by the expected cumulative future rewards. The discount factor defined by $\gamma$ controls the amount of future rewards to be taken into consideration. %With zero value of $\gamma$ the policy is completely defined by the current reward and with a value of one the policy takes into consideration future rewards upto the end of the episode.
We learn our policy using Least Square Policy Iteration (LSPI), a model-free Reinforcement Learning approach. Instead of directly estimating $Q^{\pi }(s,a)$ defined in Equation~\ref{eq_3}, the  LSPI learns the linear functional approximation of $Q^{\pi }(s,a)$ as follows:
\begin{equation}
Q^{\pi }(s,a) \approx w_{\pi}^{T}\psi(s,a)
\label{eq_6}
\end{equation} 
LSPI learns this functional approximation from the samples of MDP. Lets assume that our MDP currently is in state $\emph{s}$. It receives a reward of $\emph{r}$ by selecting an action $\emph{a}$ and transitions into state $s'$. This event will generate a sample of the form $(s, a, s', r)$ for the MDP. We generate such samples for each image until the end of an episode. We start in image as a new episode and generate such $|A|$ samples per episode. A collection of such samples define the sample set ${D}$ from which linear functional approximation weight vector $w_{\pi}$ is estimated. More specifically, lets assume that matrix ${C} \in$ $\Re ^{KxK}$ and vector ${\bf b} \in$ $\Re ^{K}$; both are initialized with matrix and vector of zeros respectively. Then ${C}$ and $b$ are updated for each consecutive samples of the form $(s, a, s', r)$ from $D$ using the  following equations\cite{Lagoudakis_JMLR2003}\cite{Kwok_IROS2004}:
\begin{equation}
{C} = {C} + \psi(s,a)(\psi(s,a)-\gamma\psi(s^{'}, \pi(s^{'}))
\label{eq_7}
\end{equation} 
\begin{equation}
{b} = {b } + \psi(s,a)r
\label{eq_8}
\end{equation}
New policy for the next iteration is computed by the following:
\begin{equation}
{w} = {C}^{-1}{b}
\label{eq_12}
\end{equation} 

%\comment{
We refer to the works of Lagoudakis et al.\cite{Lagoudakis_JMLR2003} for a detail derivation.
Starting with an initial policy, we generate a sample set ${D}$. The new policy is estimated using the equations \ref{eq_7}\ref{eq_8}. This new policy is used to generate sample set for the next iteration of the policy. We follow $\epsilon$-greedy action selection mechanism, starting with a large $\epsilon$ value that allows for sufficient space exploration of the policy during training. $\epsilon$  values are reduced in successive iterations. Although LSPI can reuse the same sample set ${D}$ generated during the first iteration, we find it more useful to use the sample sets generated in successive iterations to learn our policy. During the testing step of our policy, we set the epsilon value to be 0.005. We experimentally found that maximum of 10 iteration is sufficient for learning our policy. 
Figure ~\ref{fig:conflict}b) shows the learned weights for the \emph{living room} experiment shown in Figure\ref{fig:policy_eval_cross_val}. The weight vector is reshaped according to action blocks in each row for better demonstration.

\section{Experiments}
\subsection{ Specific Supervised Segmentation}
We conducted our experiments on the NYUD V2 dataset~\cite{Silberman_ECCV2012}. For the definition of the scene category, we followed the 9 most common scene categories as defined in~\cite{Gupta_CVPR2013}. These scenes contain multiple instances of objects and are well represented across the 27 scene categories provided in the original dataset. %We constructed the co-occurrence matrix of the 34 most frequent objects excluding the common background classes of  \emph{wall, floor} and {\em ceiling}. 
%These 34 semantic labels are analyzed in the context of semantic segmentation task and the state-of-the-art performance in \emph{Jaccard Index} metric are reported in~\cite{Gupta_ECCV2014}. 
We use the standard train/test split 795/654 images as used in \cite{Silberman_ECCV2012, Gupta_CVPR2013}. We learn the CRF parameters $w_i$ and weights of the decision tree for Adaboost classifier from the training set. We solved Equation \ref{eq_2} for 9 different scenes for each of the object categories to get binary object/background segmentations. 
% Table \ref{table:featnyud} tabulates individual dimension of our feature used for the AdaBoost classifier. \\

\begin{figure*}[t]%[!htb]
%\begin{tabular}{c@{\hspace{0.01cm}}c@{\hspace{0.01cm}}c}
\begin{tabular}{cccccc}
%\multicolumn{3} {c} %{\specialcell{\includegraphics[trim=45 710 100 40, clip=true, width=0.45\textwidth] {figures/colorcode/rezaCVPR15c.pdf}}}\\
\centering
\includegraphics[clip,width=0.15\textwidth]{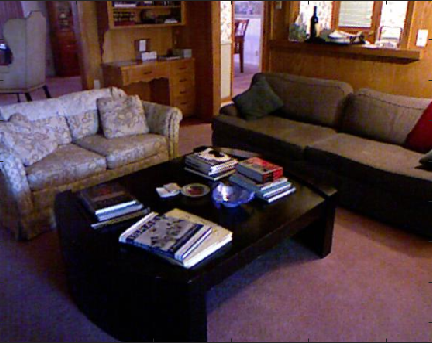}&
%\hspace{0.03cm}
\includegraphics[clip,width=0.15\textwidth]{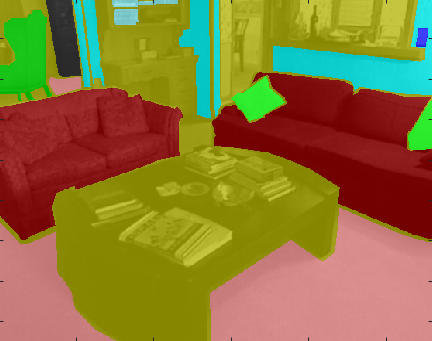}&
%\hspace{0.03cm}
\includegraphics[clip,width=0.15\textwidth]{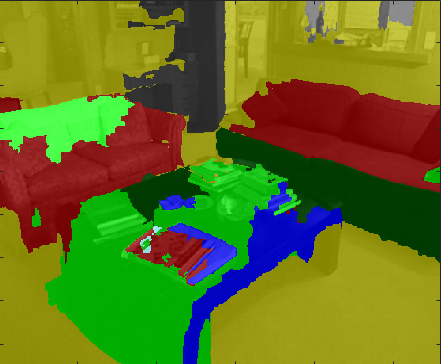}&
%\vspace{0.05cm}
\includegraphics[clip,width=0.15\textwidth]{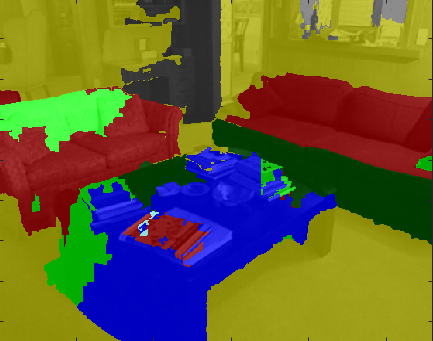}&
%\hspace{0.03cm}
\includegraphics[clip,width=0.15\textwidth]{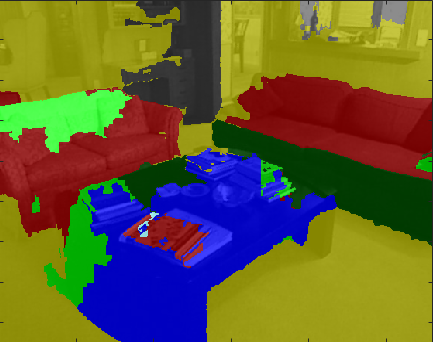}&
%\hspace{0.03cm}
\includegraphics[clip,width=0.15\textwidth]{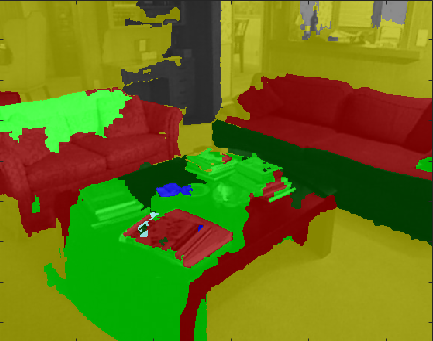}
%\vspace{0.05cm}
%\includegraphics[clip,width=0.15\textwidth]{figures/crossval/test_results_bookstore_cvfold_3.png}&
%%\hspace{0.03cm}
%\includegraphics[clip,width=0.15\textwidth]{figures/crossval/test_results_classroom_cvfold_3.png}&
%%\hspace{0.03cm}
%\includegraphics[clip,width=0.15\textwidth]{figures/crossval/test_results_home_office_cvfold_3.png}&
%%\hspace{0.03cm}
%\includegraphics[clip,width=0.15\textwidth]{figures/crossval/test_results_bathroom_cvfold_3.png}&
%%\hspace{0.03cm}
%\includegraphics[clip,width=0.15\textwidth]{figures/crossval/test_results_dining_room_cvfold_3.png}&
%%\hspace{0.03cm}
%\includegraphics[clip,width=0.15\textwidth]{figures/crossval/test_results_office_cvfold_3.png}\\
\end{tabular}
\caption{Visualization of semantic segmentations from the \emph{LSPI}-learned policy against the other baselines in \emph{livingroom} scene. From left to right the images are \emph{rgb, ground truth} following up with semantic segmentation output from \emph{LSPI, Fixed order, Random order}, and \emph{Oracle} respectively.}
\label{fig:policy_image_comparison}
\end{figure*}
\subsection{Policy Evaluation using 3-Fold Cross-Validation}
\label{sec:eval_cross_val}
\begin{figure*}[t]%[!htb]
%\begin{tabular}{c@{\hspace{0.01cm}}c@{\hspace{0.01cm}}c}
\begin{tabular}{ccc}
%\multicolumn{3} {c} %{\specialcell{\includegraphics[trim=45 710 100 40, clip=true, width=0.45\textwidth] {figures/colorcode/rezaCVPR15c.pdf}}}\\
\centering
\includegraphics[clip,width=0.30\textwidth]{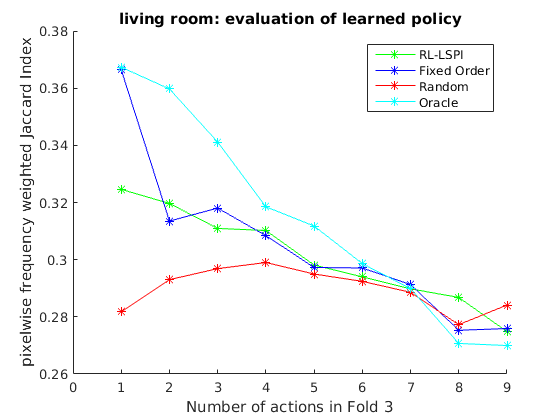}&
%\hspace{0.03cm}
\includegraphics[clip,width=0.30\textwidth]{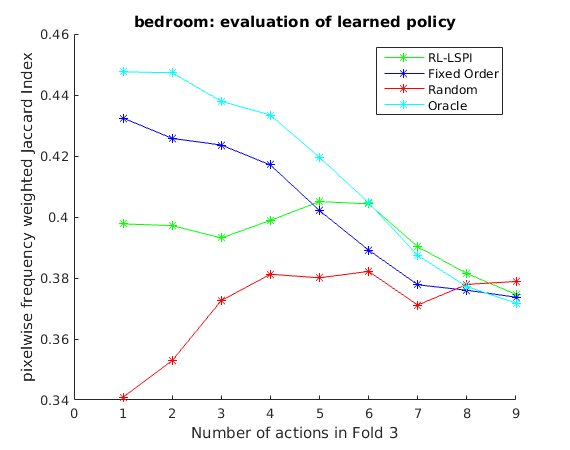}&
%\hspace{0.03cm}
\includegraphics[clip,width=0.30\textwidth]{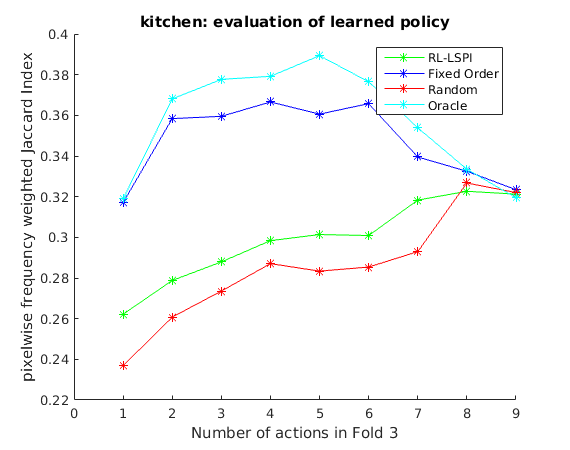}\\
%\vspace{0.05cm}
\includegraphics[clip,width=0.30\textwidth]{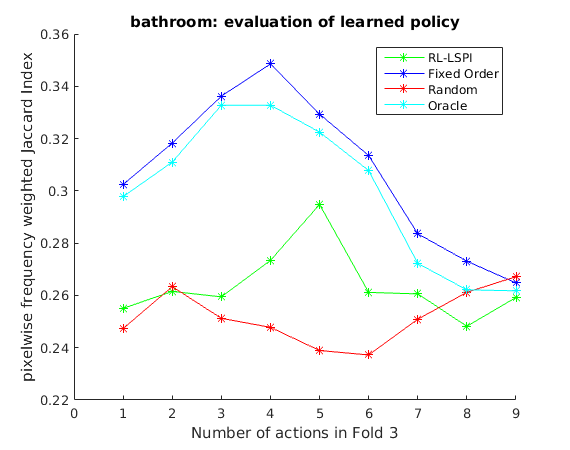}&
%\hspace{0.03cm}
\includegraphics[clip,width=0.30\textwidth]{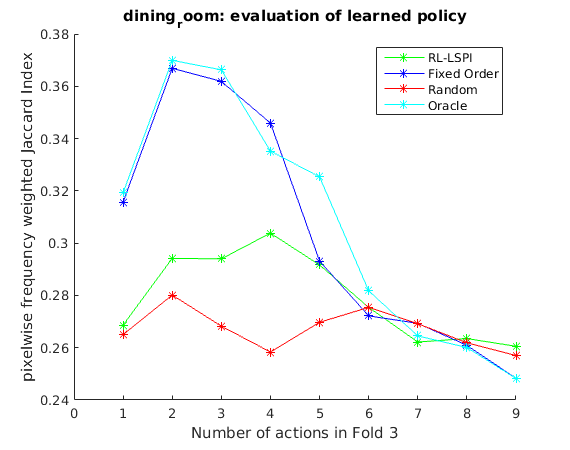}&
%\hspace{0.03cm}
\includegraphics[clip,width=0.30\textwidth]{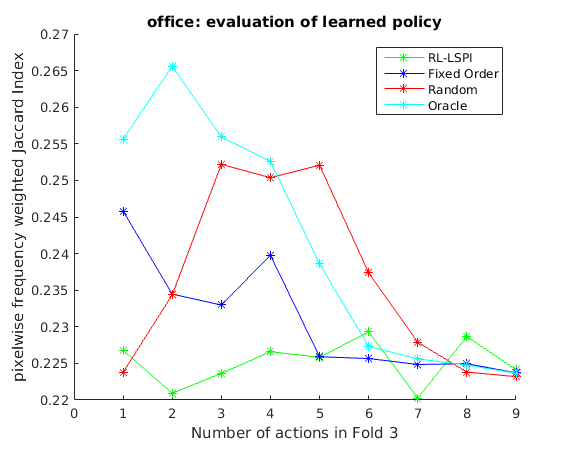}\\
%\vspace{0.05cm}
\includegraphics[clip,width=0.30\textwidth]{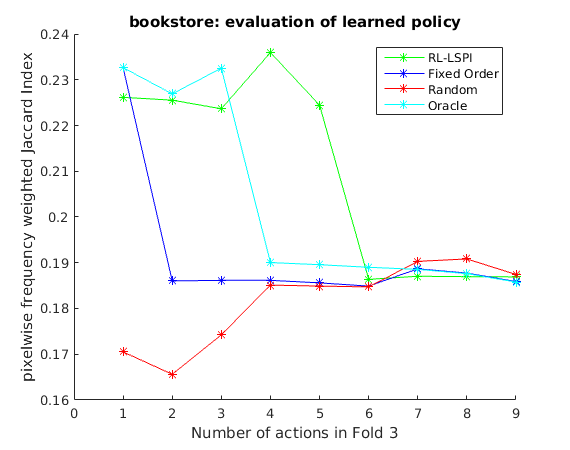}&
%\hspace{0.03cm}
\includegraphics[clip,width=0.30\textwidth]{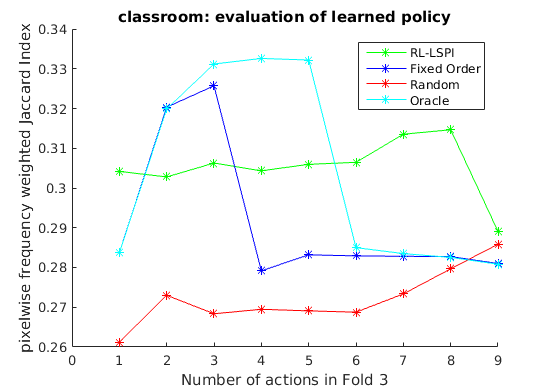}&
%\hspace{0.03cm}
\includegraphics[clip,width=0.30\textwidth]{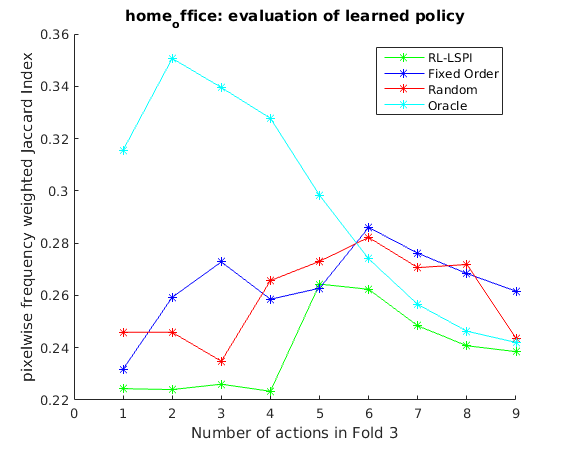}\\
\end{tabular}
\caption{Comparison (better seen in color) of our \emph{\bf LSPI} learned policies against three baselines: \emph{\bf Fixed Order}, \emph{\bf Random Order}, \emph{\bf Oracle} from 9 different scenes in NYUD-V2 dataset. Each plot shows number of actions taken into consideration in the horizontal axis. The vertical axis shows the average value of the reward metric \emph{pixelwise frequency weighted Jaccard Index} on the test-fold images. Each curve demonstrates how the policy evolves as we take more actions into consideration. The first row shows comparisons in \emph{living room, bedroom, kitchen} scenes respectively. These three scenes have the reasonably good number of images to train the policies from. Second row shows the comparisons on \emph{bathroom, dining room, office} scenes respectively. These has moderate number of images to learn the policy. The last row shows the comparisons on \emph{bookstore, classroom, home office} respectively with fewer images for policy learning.%Scene specific images used to learn our policy. %are listed in Table\ref{tab:nyud_v2_scene_spec_stats}. Here we showed the graphical comparison only on one of three folds of cross validation. The remaining two folds are depicted in the supplementary material.}
}
\label{fig:policy_eval_cross_val}
\end{figure*}

We aim at learning a dynamic policy that can effectively give us an object selection mechanism learned from the data. Learned policy should select object based on the content of the image to maximize the reward, which we define to be \emph{pixel wise frequency weighted Jaccard Index} of the image. To test our hypothesis we conducted experiment on the standard test split portion of NYUD V2 dataset, which comprises 654 images of various scenes. Note that the 795 training images of the standard split was only used to generate our supervised segmentation. %Table\ref{tab:nyud_v2_scene_spec_stats} shows the objects considered. The last column shows total number of images in each scene, where the policy is evaluated. 
We conducted a scene-specific experiment on the 9 scenes in NYUD V2. In each scene we make a 3-fold cross validation images. Train the policy using 2-fold images and test the policy on the third fold. Figure\ref{fig:policy_eval_cross_val} compares our {\em LSPI} learned policy against three other baseline policies defined as follows: \\
\noindent
i) \emph{Fixed Order Policy} In this policy mechanism, the order of selecting objects is fixed and is defined to be the most-frequently occurring 9 objects that are present in the standard split of the training images\cite{Silberman_ECCV2012}. For example, in \emph{bedroom} scene the most-frequently occurring objects under consideration are \{\emph{bed, pillow, lamp, nightstd, dresser, box, clothes, chair, books}\} and there are 191 images in the standard test split of NYUD V2. Similar statistics  are  computed  for  additional  8  scenes of \{\emph{bookstore, classroom, dinning room, homeoffce, kitchen, living-room, office}\}.\\
ii) \emph{Random Order Policy} An action is selected randomly based on the uniform distribution. The set of actions are again chosen to be the most-frequently occurring 9 objects as described in the \emph{Fixed Order Policy}. \\
iii) \emph{Oracle Policy} In this policy, an oracle tells the policy to pick from the objects that are already present in the current image. Again the set of actions are the most frequent 9 objects like the preceding two baseline policies.\\
They are colored \emph{blue, red, cyan}  respectively in the evaluation plots. Figure~\ref{fig:policy_eval_cross_val} reveals that our policy does better than \emph{Random Order Policy}  scenes with sufficient images to be trained the policy. It is comparable with the \emph{Fixed Order Policy} in most of the 9 scene experiments. The \emph{Oracle Policy} does better than other policies since it has the correct information about the presence or absence of objects in the image.

\subsection{ Policy Evaluation on Controlled Set of Images}
In the 3-Fold cross validation experiment, frequent occurring objects may dominate the test-fold images where the policy is evaluated. Hence it may cause the \emph{fixed} order policy to perform better than others in our metric of comparison. We created a more controlled set of train/test-fold images to evaluate the learned policy. In this experiment, we consider the most-frequently occurring 9 objects in \emph{living room} scene from the image in standard test split of NYUD-V2 dataset. These are {\em sofa, table, pillow, chair, lamp, cabinet, books, bookshelf} and {\em bag}. We create two control sets by deliberately choosing those images for the train/test fold, where a pair of most-frequently occurring objects are does not appear in every image of the folds. In \emph{Control set I}, the pair chosen is \emph{pillow, chair} and in \emph{Control set II} the objects chosen are \emph{table, chair}.

\begin{figure*}[t]%[!htb]
%\begin{tabular}{c@{\hspace{0.01cm}}c@{\hspace{0.01cm}}c}
\begin{tabular}{ccc}
%\multicolumn{3} {c} %{\specialcell{\includegraphics[trim=45 710 100 40, clip=true, width=0.45\textwidth] {figures/colorcode/rezaCVPR15c.pdf}}}\\
\centering
\includegraphics[clip,width=0.30\textwidth]{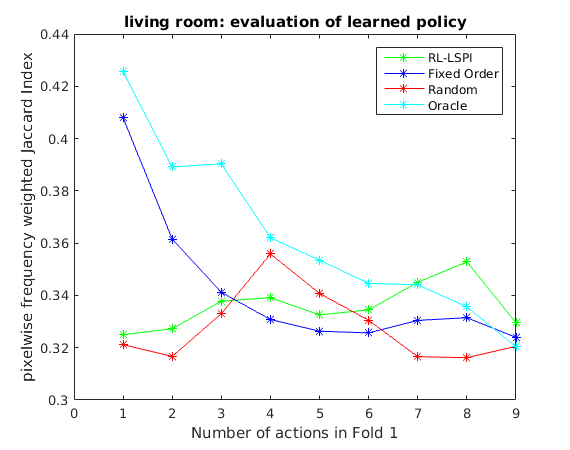}&
%\hspace{0.03cm}
\includegraphics[clip,width=0.30\textwidth]{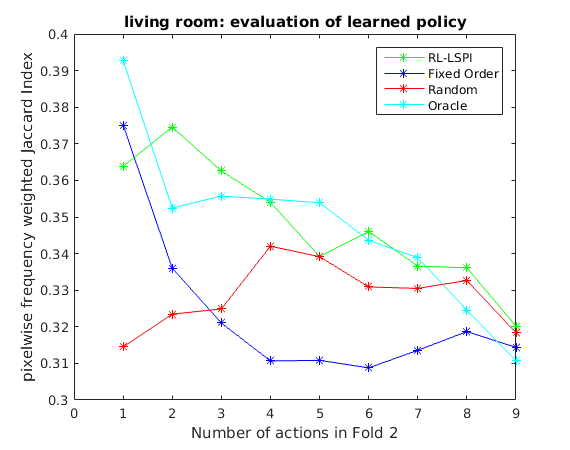}&
%\hspace{0.03cm}
\includegraphics[clip,width=0.30\textwidth]{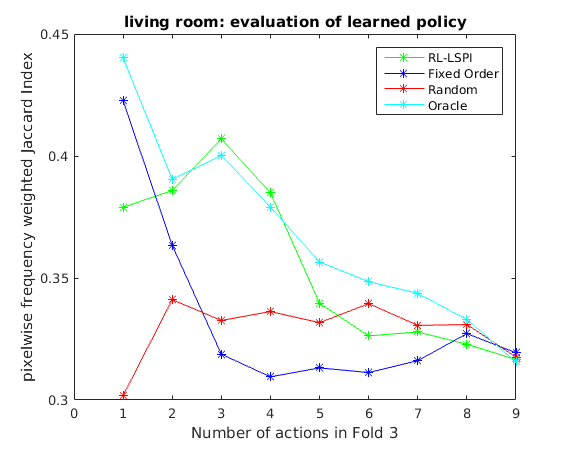}\\
%\vspace{0.05cm}
\includegraphics[clip,width=0.30\textwidth]{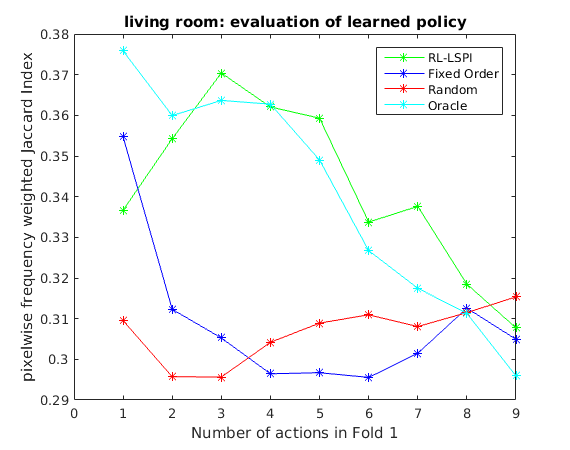}&
%\hspace{0.03cm}
\includegraphics[clip,width=0.30\textwidth]{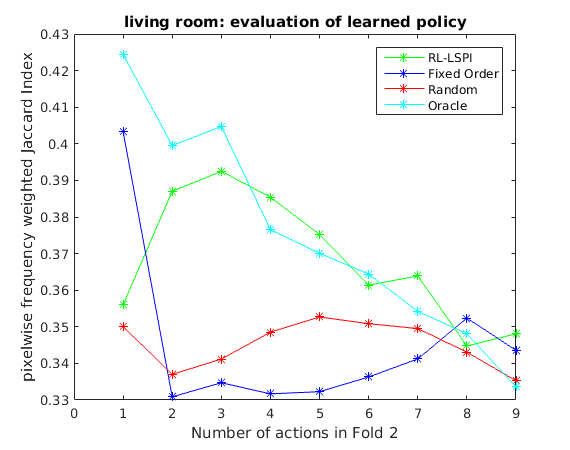}&
%\hspace{0.03cm}
\includegraphics[clip,width=0.30\textwidth]{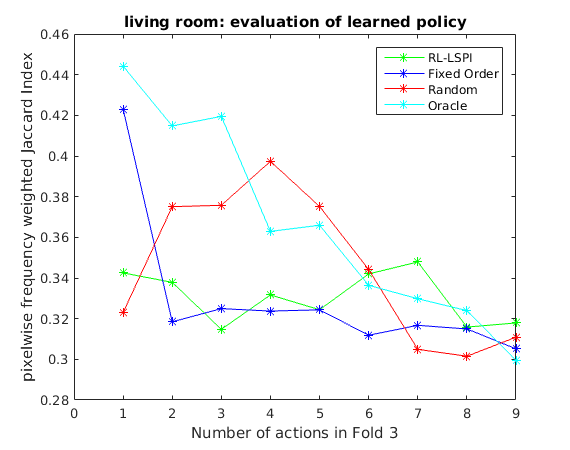}\\
\end{tabular}
\caption{Comparison of our \emph{LSPI}-learned policies against three baselines: \emph{Fixed Order}, \emph{Random Order}, \emph{ Oracle} in the \emph{living room} scene in NYUD-V2 dataset. Each plot shows number of actions taken into consideration in the horizontal axis. The vertical axis shows the reward metric, i.e., average of the \emph{pixelwise frequency weighted Jaccard Index} on the test-fold images. The first row shows comparisons in \emph{Control set I}. Second row shows the comparisons in \emph{Control set II}. Notice in this control experiment, a pair of frequently occurring objects are by deliberately choosing those images for the train/test fold, where a pair of most-frequently occurring objects are does not appear in every image of the folds. Our policy performs better than both \emph{Random, Fixed} order policies and comparable to \emph{Oracle}}
\label{fig:policy_eval_control}
\end{figure*}

% \footnote{Notice in the previous experiment, the frequency statistics are found from the standard train split of NYUD-V2 dataset}. The objects considered are tabulated in Table\ref{tab:nyud_v2_control_exp_stats}. 
A similar graphical comparison (as in previous experiment) of our learned policy with the three baselines are shown in Figure~\ref{fig:policy_eval_control}.

%\comment{
%% \newcommand{\rtt}[1]{\begin{sideways}{#1}\end{sideways}}
%\begin{table}[h]
%\footnotesize
%%\begin{center}
%\begin{tabular}{|p{5.5em}|p{1.0em}|p{1.0em}|p{1.0em}|p{1.0em}|p{1.0em}|p{1.0em}|p{1.0em}|p{1.0em}|p{1.0em}|}
%\hline
%\ Livingroom        &\rtt{sofa}& \rtt{table}& \rtt{pillow}& \rtt{chair}& \rtt{lamp}& \rtt{cabinet}& \rtt{books}& \rtt{booksh}& \rtt{bag}\\
%%\ Livingroom        &\rtt{sofa}& \rtt{table}& \rtt{chair}& \rtt{cabinet}& \rtt{pillow}& \rtt{lamp}& \rtt{bag}& \rtt{tv}& \rtt{books}\\
%\hline
% \end{tabular}
%%\end{center}
%\caption{Most frequent 9 objects in \emph{living room} considered in the control experiment.}
%\label{tab:nyud_v2_control_exp_stats}
%\end{table}
%}

\begin{table}[ht]
 \footnotesize

 %\begin{center}

   \begin{tabular}{p{3.0em}|p{1.1em}|p{1.1em}|p{1.1em}|p{1.1em}|p{1.1em}|p{1.1em}|p{1.1em}|p{1.1em}|p{1.1em}|p{1.1em}|p{1.1em}|p{1.1em}|p{1.1em}|p{1.1em}|p{1.1em}|p{1.1em}|p{1.1em}|p{1.1em}|p{1.1em}}

  &\rtt{Bedroom} & \rtt{Bathroom}  & \rtt{Bookstore}  & \rtt{Diningroom}  & \rtt{HomeOffice} & \rtt{Kitchen} & \rtt{Livingroom}  & \rtt{Office}  & \rtt{Classroom}  & \rtt{Gupta\cite{Gupta_CVPR2013}} & \rtt{Gupta\cite{Gupta_ECCV2014}} \\

 \hline

 \ bed           	& 41.5 & - & - & - & - & - & - & - & - & 57.0 & 60.5 \\                      

 \ pillow           & 20.3 & - & - & - & - & - & 13.5 & - & - & 30.3 & 34.4 \\                      

 \ lamp         	& 9.5  & - & 0 & 5.6 & 0.43 & - & 6.4 & - & - &  16.3&  34.8\\          

 \ nt-std      & 8.5  & - & - & - & - & - & - & - & - &  21.5&  27.2\\                      

 \ dressr        	& 15.1 & - & - & - & - & - & - & - & - & 24.3 & 34.8 \\                      

 \ box           	& 1.7  & - & 0 & - & - & 3.0 & - & 8.7 & 3.0 &  2.1 & 2.1 \\                      

 \ cloth        	& 5.7  & - & 6.0 & - & - & - & - & - & - &  7.4 & 4.7 \\                      

 \ chair            & 13.0 & - & - & 29.0 & 12.0 & 19.3 & 10.6 & 32.6 & 24.5 & 36.7 & 47.9 \\ 

 \ books         	& 6.6  & - & 5.9 & - & 13.7 & - & 2.6 & 3.1 & 0 & 5.5 & 6.4 \\

 \ wall 			& 56.3 & 37.8 & 16.4 & 36.5 & 45.9 & 35.5 & 41.9 & 35.3 & 26.7 & 67.6 & 68.0 \\

 \ floor			& 73.9 & 43.9 & 58.5 & 66.9 & 67.1 & 79.1 & 62.6 & 76.8 & 56.3 & 81.2 & 81.3 \\

 \ ceil 			& 32.0 & 0 & 17.9 & 55.4 & 0 & 23.2 & 28.4 & 38.9 & 44.6 & 61.1 & 60.5 \\

% \ background		& 16.4 & 9.9 & 11.0 & 15.0 & 19.2 & 19.5 & 16.7 & 15.1 & 24.2 &  &  \\

 \ bkshf		    & - & - & 13.8 & - & 14.0 & - & - & - & - &  19.5 & 18.1 \\

 \ shelv			& - & 5.5 & 6.4 & 0.08 & - & - & - & 2.7 & 12.1 & 4.5 & 3.5 \\

 \ cabint			& - & 24.1 & 0 & 19.4 & 5.6 & 33.8 & 5.0 & 4.6 & 9.3 & 44.8 & 44.9 \\

 \ table			& - & - & 3.9 & 17.9 & 4.1 & 9.0 & 8.7 & 7.0 & 9.5 & 28 & 29.9 \\

 \ desk				& - & - & 0 & - & 13.6 & - & - & 3.0 & 23.6 & 7.1 & 11.3 \\

 \ person			& - & - & - & - & - & - & - & - & 0 &  5 &  0.2\\

 \ bag				& - & 0.46 & - & 2.0 & 3.9 & 2.6 & 4.4 & 0 & 0 &  0 & 0.2 \\

 \ sofa				& - & - & - & 1.7 & 11.9 & - & 20.3 & - & - & 40.8 &  47.9\\

 \ countr			& - & 42.1 & - & 0.95 & - & 29.6 & - & 0.55 & - & 52 & 51.3  \\

 \ curtn			   & - & - & - & 21.8 & - & - & - & - & - & 28.6 & 29.1 \\

 \ sink				& - & 21.7 & - & - & - & 14.6 & - & - & - & 35.7 & 37.5 \\

 \ fridge 	& - & - & - & - & - & 16.7 & - & - & - & 16.2 & 14.5 \\

 \ towel			& - & 7.8 & - & - & - & 6.0 & - & - & - & 25.9 & 16.3 \\

 \ tv		& - & - & - & - & - & - & 3.7 & - & - &  4.8 & 31 \\

 \ toilet			& - & 25.6 & - & - & - & - & - & - & - & 46.5 & 55.1 \\

 \ bath			& - & 19.7 & - & - & - & - & - & - & - &  31.1 &  38.2 \\
 \ showr 	    & - & 11.5 & - & - & - & - & - & - & - &  9.7 &  4.2 \\                                           

 \hline
  \end{tabular}

 \caption{Performance comparison on different scenes in NYUD-V2 dataset in pixelwise percentage Jaccard Index.}
 \label{tab:quant_comparison_nyud}
 \end{table}
\subsection{Comparison of Sequential against Optimal Combination}
\begin{figure}[h]%[!htb]
\centering
\includegraphics[clip,width=0.40\textwidth]{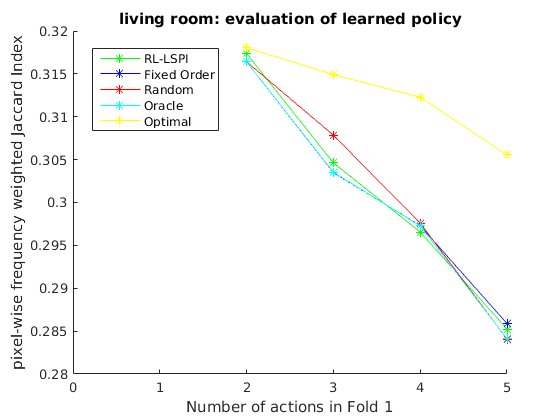}
\caption{Comparison of our \emph{LSPI} learned policies against the \emph{Optimal Combination} as well as three baselines: \emph{Fixed Order}, \emph{Random Order}, \emph{Oracle} in the \emph{living room} scene in NYUD-V2 dataset. The yellow curve shows the evolution of \emph{Optimal Combination} as more actions are taken into consideration.}
\label{fig:policy_eval_optm}
\end{figure}

We evaluated the sequential order of object selection against the optimal combination. In order to find the optimal combination of objects order, we enumerate all the permutation of $K$ objects in consideration. For each sequence of the permutation, we sequentially combine the supervised segmentations of the objects and compute the reward. Out of all such permutations, we pick the one that gives us the maximum reward as the optimal combination. The number of permutation grows exponentially with the increase of number of total objects $K$ in consideration. In this experiment, we picked 5 most frequently occurring objects in {\em living room} scene and find out the optimal combination by enumerating all 120 permutations in each image. We pick the as optimal combination the one that gives the maximum rewards in that image. Figure\ref{fig:policy_eval_optm} the graphical plot of the evaluation.

\subsection{Comparison with Other Methods}
The quantitative comparison of the semantic segmentation output from our \emph{LSPI}-learned policy on the 3-folded cross-validated experiment (explained in Section\ref{sec:eval_cross_val}) is reported in Table\ref{tab:quant_comparison_nyud}. We would like to point out that our experiments are designed in a scene specific manner and we deliberately picked a subset of all possible objects that were experimented in other methods\cite{Gupta_ECCV2014}. Here we report how our method compare against those subset of objects in each scene specific experiment. Figure\ref{fig:policy_image_comparison} demonstrates some qualitative results on \emph{living room} scene.

\section{Conclusion} 
\label{sec:conclusion}
We have demonstrated an alternative approach for semantic segmentation which affords simplicity and modularity, 
with comparable performance to the state of the art. The methodological differences come from the need to having 
adapting the existing models to different tasks, instead of training large complex models where the task is to maximize 
performance on the benchmark dataset. In the process we have also used alternative superpixel representations which favorably 
exploit both geometric and appearance properties of indoors environments. In the current work we evaluate the performance 
with respect to the task of semantic segmentation, but in the future we plan to explore the utility of the proposed representation for a large variety of tasks. 

\section*{Acknowledgments}

%% Use plainnat to work nicely with natbib. 

\bibliographystyle{plainnat}
\bibliography{references}

\end{document}